\documentclass[11pt]{article}

\usepackage{EMNLP2023}

\usepackage{times}
\usepackage{latexsym}

\usepackage[T1]{fontenc}

\usepackage[utf8]{inputenc}

\usepackage{microtype}

\usepackage{inconsolata}

\usepackage{graphicx}
\usepackage{subcaption}
\usepackage{booktabs}
\usepackage{afterpage}
\usepackage{wrapfig}

\usepackage[utf8]{inputenc}
\usepackage[hindi,english]{babel}
\def\bng{\bnwxi}
\font\bnwxi=bangwd10

\newcommand\blfootnote[1]{%
  \begingroup
  \renewcommand\thefootnote{}\footnote{#1}%
  \addtocounter{footnote}{-1}%
  \endgroup
}

\title{Crosslingual Retrieval Augmented In-context Learning for Bangla}

\author{Xiaoqian Li$^{1}$ \qquad Ercong Nie$^{1, 2}$ \qquad Sheng Liang$^{\dag}$$^{1,2}$ \\
$^{1}$Center for Information and Language Processing (CIS), LMU Munich, Germany \\
$^{2}$ Munich Center for Machine Learning (MCML), Germany \\
\texttt{Xiaoqian.Li@campus.lmu.de} \\
\texttt{\{nie, shengliang\}@cis.lmu.de}}

\begin{document}
\maketitle

\begin{abstract}
The promise of Large Language Models (LLMs) in Natural Language Processing has often been overshadowed by their limited performance in low-resource languages such as Bangla. To address this, our paper presents a pioneering approach that utilizes cross-lingual retrieval augmented in-context learning. By strategically sourcing semantically similar prompts from high-resource language, we enable multilingual pretrained language models (MPLMs), especially the generative model BLOOMZ, to successfully boost performance on Bangla tasks. Our extensive evaluation highlights that the cross-lingual retrieval augmented prompts bring steady improvements to MPLMs over the zero-shot performance. 
\blfootnote{$^\dag$ Corresponding author.}

\end{abstract}

\section{Introduction}
In recent years, the field of Natural Language Processing (NLP) has witnessed transformative advancements, especially with the advent of deep transformer techniques~\citep{Vaswani2017AttentionIA, devlin-etal-2019-bert, Radford2019LanguageMAGPT}. The introduction of Large Language Models (LLMs), such as GPT-3~\citep{Brown2020LanguageMA} and GPT-4~\citep{openai2023gpt4}, has further revolutionized the landscape. These models showcase unparalleled prowess in tasks like text classification and generation, unified under the umbrella of in-context learning, and cater to a plethora of applications across diverse languages~\citep{conneau-etal-2020-unsupervised, raffel2020exploring, Radford2019LanguageMAGPT}. While comprehensive benchmarks like XTREME~\citep{Hu2020XTREMEAM} and BUFFET~\citep{asai2023buffet} underscore their capabilities, languages such as English remain the primary beneficiaries. In stark contrast, several low-resource languages, Bangla being a prime example, grapple with challenges, notably the scarcity of pretraining corpora~\citep{artetxe-schwenk-2019-massively, hangya-etal-2022-improving,sazzed-2020-cross}.

\begin{figure}
    \centering
    \includegraphics[width=1\linewidth]{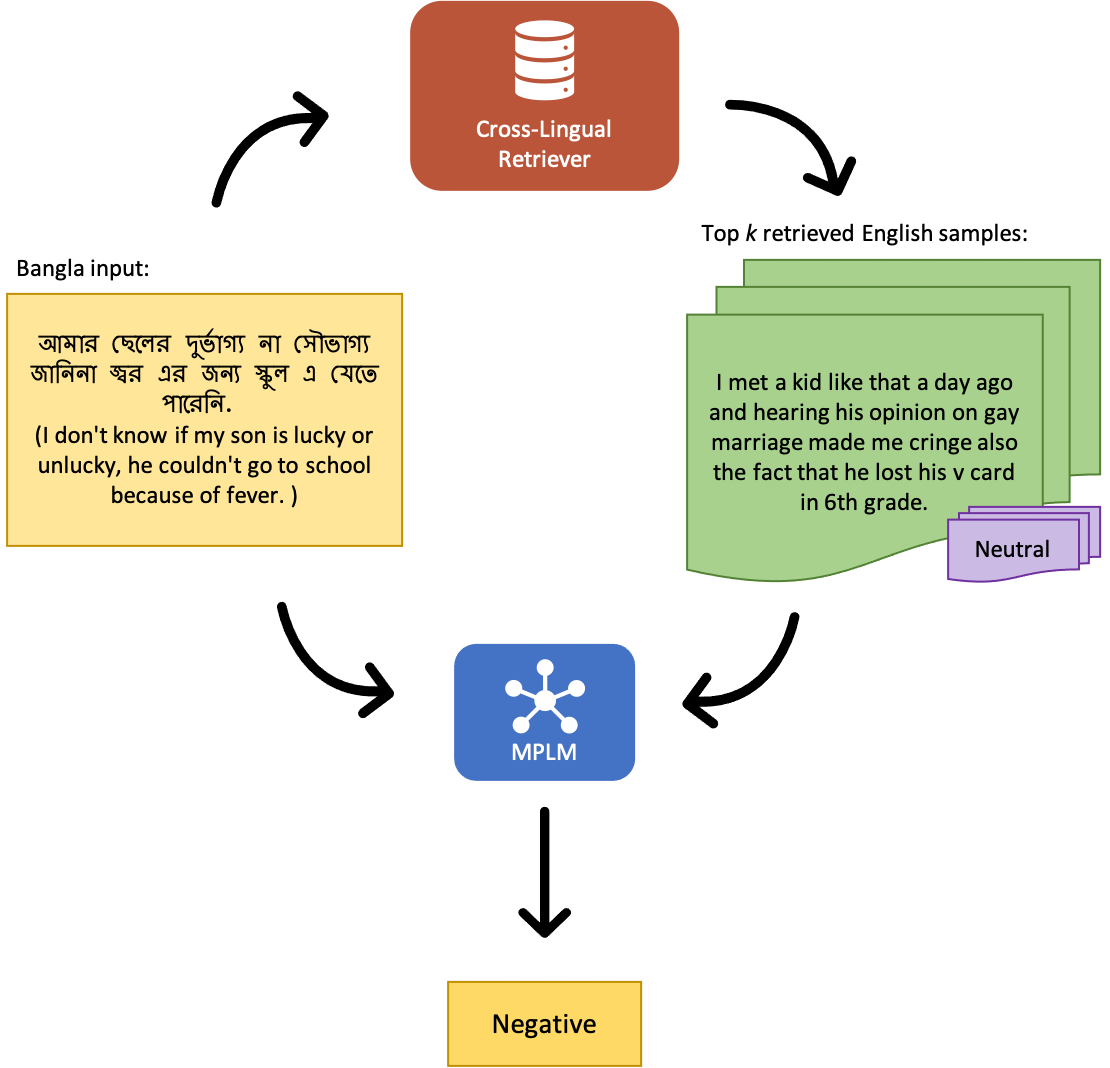}
    \caption{PARC pipeline using decoder-only Multilingual Pretrained Language Models.}
    \label{fig:parc_pipeline}
\end{figure}

Despite having a significant number of native speakers, Bangla remains underrepresented in the NLP arena due to linguistic intricacies, limited labeled datasets, and prevalent issues like data duplication~\citep{das2010phrase, das-gamback-2014-identifying}. Although there have been commendable strides using conventional machine learning techniques in Bangla NLP tasks, the untapped potential of the latest LLMs is evident~\citep{bhowmick-jana-2021-sentiment,wahid2019cricket, Hoq2021SentimentAO}.

In the evolving landscape of in-context learning with LLMs, the concept of retrieval augmentation, which emphasizes sourcing semantically rich prompts, has gained traction~\citep{shi2023replug}. However, when it comes to multilingual in-context learning, previous works like MEGA~\citep{ahuja2023mega} often limit their scope to task instructions and lack deeper semantic insights due to their approach of random prompt selection.  In contrast, strategies like PARC~\citep{nie-etal-2023-cross} pave the way for a more comprehensive methodology, fetching semantically aligned prompts from high-resource languages.

Our work draws inspiration from these methodologies but introduces novel perspectives. While MEGA offers task-level instructions, we infuse semantic understanding into our approach. Similar to PARC, our approach is cross-lingual, ensuring a broader application spectrum. Diverging from PARC's focus on masked language models like mBERT and XLMR, as shown in Figure~\ref{fig:parc_pipeline}, we venture into uncharted territories by employing larger, decoder-only multilingual pretrained language models (MPLMs) — BLOOM and BLOOMZ — to tackle Bangla NLP tasks in a generative style~\citep{muennighoff-etal-2023-crosslingual, Scao2022BLOOMA1}.

In this paper, we explore the application of cross-lingual retrieval augmented in-context learning to Bangla text classification and summarization tasks. Our main contributions encompass:
\begin{itemize}
    \item An extensive evaluation of cross-language retrieval augmented in-context learning methods in Bangla, achieving steady improvements over the zero-shot performance of MPLMs.
    \item A pioneering exploration to extend PARC to the generative models, BLOOM and BLOOMZ, providing insights for a unified pipeline of cross-lingual retrieval augmented in-context learning.
\end{itemize}

\section{Related Work}




\begin{figure*}[t]
	\centering
	\includegraphics[width=1\linewidth]{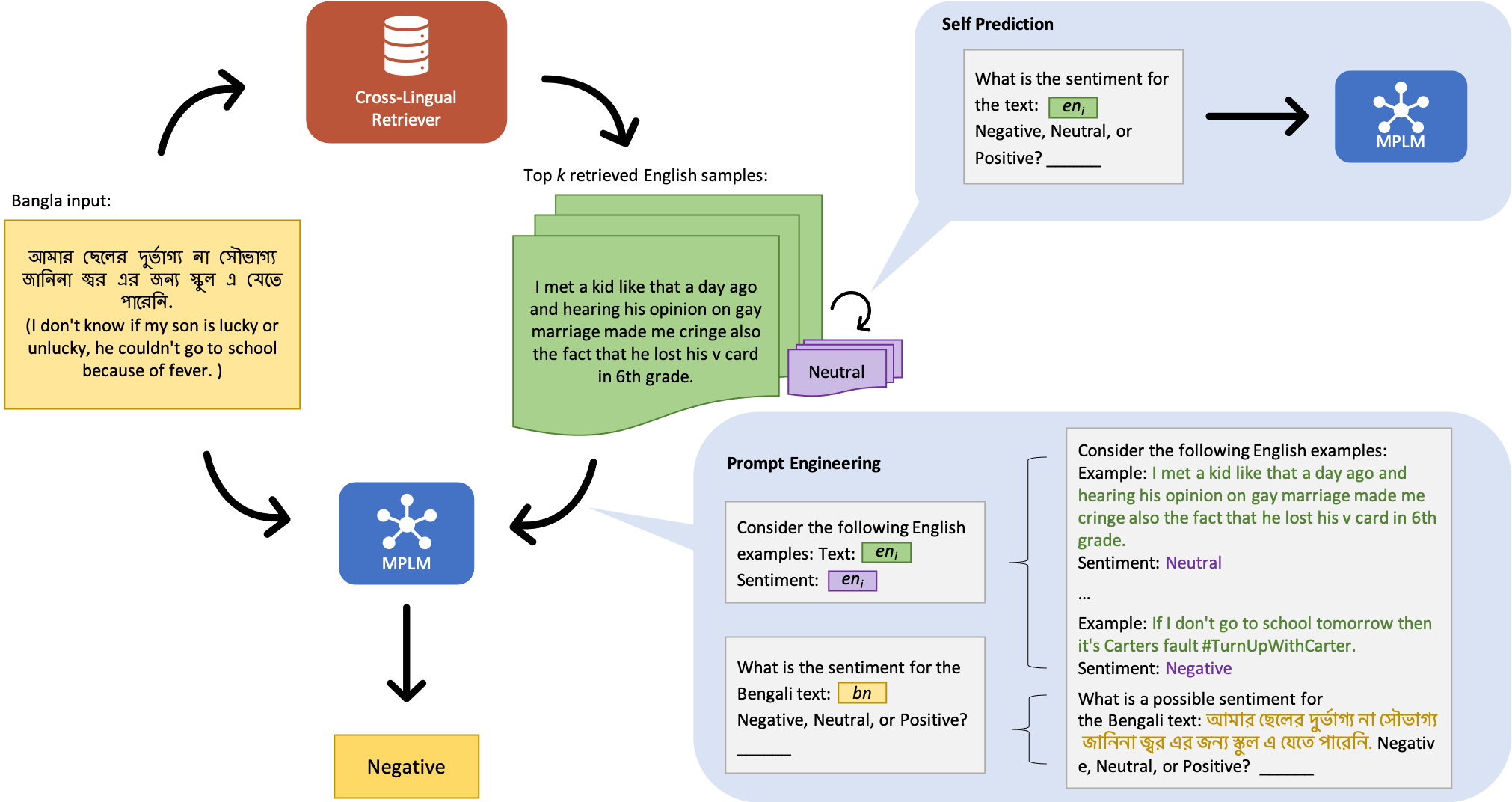}
	\caption{Detailed overview of the PARC pipeline for LRLs using cross-lingual retrieval: (a) An LRL input is used as a query for the cross-lingual retriever, which then retrieves the most semantically similar HRL sample from the HRL corpus. The associated label is either taken directly from the corpus (labeled setting) or determined by self-prediction (unlabeled setting). (b) Next, this HRL sample, its label, and the original input are combined to create a retrieval-enhanced prompt for MPLM prediction.}
	\label{fig:parc}
\end{figure*}

\paragraph{Bangla Natural Language Processing}
Bangla is a morphologically rich language with various dialects that belongs to the Indo-Aryan branch of the Indo-European language family. With roughly 270 million speakers concentrating in Bangladesh and some regions of India, Bangla is ranked as the 7th most widely spoken language in the world\footnote{\url{https://www.ethnologue.com/insights/ethnologue200/}}. However, Bangla is still considered as a low-resource language in the NLP research due to the scarcity of digital text resources and annotated corpora. 

Research on Bangla NLP has covered a variety of common NLP subfields since 1990s, such as POS tagging~\citep{dandapat2004hybrid, ekbal2008web}, stemming and lemmatization~\citep{islam2007light, paik2008simple}, named entity recognition~\citep{ekbal2007hidden, ekbal2008development}, sentiment analysis~\citep{das2010phrase, wahid2019cricket}, news categorization~\citep{mansur2006analysis, mandal2014supervised}, etc. However, the research in different areas of Bangla NLP still remains sparse.
In the era of deep learning, further progress has been made in Bangla NLP, particularly in terms of the development datasets~\citep{rahman2018datasets, islam-etal-2021-sentnob-dataset, islam2023sentigold} and models~\citep{tripto2018detecting, ashik2019data, karim2020classification}. Pretrained language models have achieved decent performance in a large variety of NLP downstream tasks through the finetuning. Under this background,~\citet{bhattacharjee-etal-2022-banglabert} pretrained the BanglaBERT model, a BERT-based language understanding model pretrained on Bangla language corpora. With the advent of the large language models (LLMs), zero- and few-shot prompting methods have gradually gained prominence.~\citet{hasan2023zero} compared the zero- and few-shot prompting performance of LLMs with the finetuned models for the Bangla sentiment analysis task. Our work explores the application of the retrieval-augmented prompting method in Bangla violence detection and sentiment analysis tasks.

\paragraph{Multilingual In-context Learning}

\citet{brown2020language} demonstrated that LLMs like GPT-3 can acquire task-solving abilities by incorporating input-output pairs as context. The in-context learning approach involves concatenating input with randomly selected examples from the training dataset, which is also called the prompting method. Recent research \citep{gao-etal-2021-making, liu-etal-2022-makes, liu-etal-2023-semantic, shi2023replug} has expanded on this idea by enhancing prompts for pretrained models through the inclusion of semantically similar examples. 
The effectiveness of prompting methods for English models extends to multilingual models in cross-lingual transfer learning as well. \citet{zhao-schutze-2021-discrete} and \citet{huang-etal-2022-zero} investigated the prompt-based learning with multilingual PLMs.~\citet{nie-etal-2023-cross} incorporated augmented the prompt with cross-lingual retrieval samples in the multilingual understanding and proposed the PARC pipeline.~\citet{tanwar-etal-2023-multilingual} augmented the prompt with not only cross-lingual semantic information but also additional task information. However, previous studies mainly concentrated on the multilingual encoder or encode-decoder models, while our work extend the PARC pipeline to the decoder-only multilingual LLMs.

\paragraph{Multilingual LLMs}
In the era of LLMs, BLOOMZ and mT0~\citep{muennighoff-etal-2023-crosslingual} are two representative newly emerging multilingual models. These two multilingual LLMs are finetuned on xP3, a multilingual multitask finetuning dataset, and based on the pretrained models BLOOM~\citep{Scao2022BLOOMA1} and mT5~\citep{xue-etal-2021-mt5}, respectively. Six different sizes of BLOOMZ models are released from 560M to 176B and 5 different sizes of mT0 models are released from 300M to 13B. These multilingual LLMs open up the possibility for conducting few- and zero-shot cross-lingual in-context learning, as demonstrated by recent benchmarking efforts, for example MEGA~\citep{ahuja2023mega} and BUFFET~\citep{asai2023buffet}.

\section{Methodology}

Our research extends the work of \citet{nie-etal-2023-cross} by focusing on improving multilingual pre-trained language models (MPLMs) for low-resource languages in a zero-shot setting, specifically using retrieved content from high-resource languages such as English.

The backbone of our research approach is a two-stage pipeline consisting of a cross-lingual retriever and a prompt engineering process as shown in Figure~\ref{fig:parc}. 
This pipeline aims to build on the strengths of MPLMs while mitigating their limitations, especially when dealing with low-resource languages. 
The first stage of the pipeline uses a cross-lingual retriever that maps the input Bangla text $q$ to a vector $q_{embed}$ in a shared embedding space and uses it as a query. 
Using semantic similarities with $q_{embed}$, the retriever returns the most similar k examples from high-resource languages either with or without their labels:
$$R = \arg\max_{i \in \{1, \ldots, |d|\}}^k \cos(q_{embed}, d_i)$$
where \( d_i \) means each document in the high-resource language corpus and \( |d| \) is the number of documents. If there's no label, it suggests a self-prediction step.

The second stage of the pipeline is the prompt engineering. The input Bangla text and the retrieved pattern are subjected to this process. A prefix prompt template \( P \) is used to reformulate the input to facilitate the model's prediction \( \mathbf{y} \):
$$\mathbf{y} = MPLM( P(q, R))$$
Depending on the architecture of the chosen MPLM, for decoder-only models, the answer is generated by the model directly. 
For encoder models, the answer is obtained by first mapping each label to its predefined word using the \textit{verbalizer} and then deducing the label word using mask token prediction.

By integrating cross-lingual content retrieval with prompt-guided prediction, we aim to improve the ability of MPLMs to handle low-resource languages. This synergy not only extracts rich linguistic insights from high-resource languages, but also uses them to improve performance on low-resource language tasks.

\section{Experiments}
In this study, we focused on the tasks of classification and summarization. We refer to our research approach, which uses k retrieved samples for cross-lingual augmented in-context learning methods, as the main method in the following sections.

\subsection{Baselines}
\paragraph{Zero-shot}The template, when populated with the input sample, is fed directly into the MPLM for prediction. This process bypasses the use of cross-lingual context. 
\paragraph{Lead64} The first 64 tokens of the input text are taken as a summary of the text (For summarization tasks only).

\subsection{Tasks}
\subsubsection{Classification}
\paragraph{Vio-Lens} The Vio-Lens dataset~\citep{SahaAndJunaed} contains YouTube comments related to violent incidents in the Bengal region, with the goal of highlighting potential threats that could incite further violence.
The prompt templates for both main method and zero-shot baseline are defined as follows:

\begin{itemize}
	\item BLOOMZ-3b and BLOOM-3b: \\
 \texttt{Reflecting on the statement "\{text\}", which aggressive level does it resonate with: non-aggressive, slightly aggressive, or highly aggressive?}
	\item mBERT:
 \texttt{The underlying theme in \{text\} is {[}MASK{]}.} \\
 with the verbalizer:\\
 $v(\textit{Direct Violence})= $ \texttt{assaultive}, \\
 $v(\textit{Passive Violence}) = $ \texttt{indirect}, \\
 $v(\textit{Non-Violence})= $ 
  \texttt{peaceful}

\end{itemize}

We use the ETHOS (onlinE haTe speecH detectiON dataSet)~\citep{mollas2020ethos} as sentence pool in our experiments. This repository provides a dataset designed to identify hate speech on social media. We use the binary variants of the dataset, which contains 998 comments, each labeled for the presence or absence of hate speech. Since the labels are inconsistent, we use the self-prediction method to predict the labels.

\paragraph{SentNoB} Designed to capture the sentiment within text, SentNoB classifies content as positive, negative or neutral~\citep{islam-etal-2021-sentnob-dataset}.
The prompt templates for both main method and zero-shot baseline are defined as follows:
\begin{itemize}
	\item BLOOMZ-3b and BLOOM-3b:\\ \texttt{Text: \{text\} What is a possible sentiment for the text given the following options?}
	\item mBERT: \texttt{\{text\} Sentiment: {[}MASK{]}}  \\
    with the verbalizer: \\
    $v(0) = $ \texttt{positive},  $v(1) = $ \texttt{neural}, \\
    $v(2) = $ \texttt{negative}
\end{itemize}

The English Sentiment Analysis dataset~\citep{rosenthal-etal-2017-semeval}, which consists of tweets annotated for sentiment on 2-, 3-, and 5-point scales with labels positive, negative, and neutral, serves as the HRL corpora in our study. We use the labeled training set for our experimental sentence pool.

\subsubsection{Summarization}

\paragraph{XL-Sum}
is a large and varied dataset consisting of 1.35 million pairs of articles and their corresponding summaries~\citep{hasan-etal-2021-xl}. These pairs have been expertly annotated by the BBC and meticulously extracted through a series of carefully designed heuristic methods. The dataset includes 45 languages, from low to high resource, many of which do not currently have publicly available datasets.
The prompt template is defined for all models as follows:
\begin{itemize}
	\item Main method: \\
 \texttt{\{text\} Generate a concise summary of the above text using the same language as the original text (\{target\_lang\})}:
        \item Zero-shot baseline: \\
        \texttt{\{text\} Generate a concise summary of the given text:}  
\end{itemize}

\subsection{Models}

\paragraph{BLOOM} is an autoregressive Large Language Model trained on a diverse corpus to generate text based on prompts~\citep{Scao2022BLOOMA1}. It is capable of generating coherent text in 46 languages. 

\paragraph{BLOOMZ} takes a novel approach in the MPLM landscape by applying Bloom filters in the context of language models~\citep{muennighoff-etal-2023-crosslingual}. This allows the model to use high-resource languages to improve embeddings for low-resource languages, effectively bridging the gap between languages with different levels of available resources. 

\paragraph{mBERT} is an early MPLM that extends the original BERT model~\citep{DBLP:journals/corr/abs-1810-04805}. It is pre-trained on a corpus of 104 languages, using shared WordPiece vocabularies and a unified architecture for all languages.

\paragraph{mT5} or Multilingual T5~\citep{xue-etal-2021-mt5}, is an extension of the T5 (Text-to-Text Transfer Transformer) model~\citep{raffel2020exploring} designed specifically for multilingual capabilities. Pre-trained on mC4, a large multilingual dataset, mT5 demonstrates multilingual capabilities by transforming input text sequences into output sequences. 


\paragraph{Cross-Lingual Retriever} We followed  \citet{nie-etal-2023-cross} to use 
the multilingual sentence transformer ``\textit{paraphrase-multilingual-mpnet-base-v2}''~\citep{reimers-gurevych-2019-sentence}. This transformer maps sentences and paragraphs into a 768-dimensional dense vector space. Such a high-dimensional embedding facilitates tasks such as clustering and semantic search.
In our experiments, the number of retrieval samples $k$ is 1 and 3 for classification task and 1 for summarization task.

\section{Results}
\subsection{Results of classification tasks}

\begin{table}[]
\footnotesize
\centering \begin{tabular}{lccc}
\toprule
Vio-Lens          & zero shot & k=1  & k=3  \\ 
\midrule
bloomz-3b & 0.19      & 0.2  & 0.24 \\ 
bloom-3b  & 0.00      & 0.00 & 0.00 \\ 
mbert     & 0.21      & 0.28 & 0.29 \\ 
\midrule
SentNoB          & zero shot & k=1  & k=3  \\
\midrule
bloomz-3b & 0.34      & 0.44 & 0.44 \\ 
bloom-3b     & 0.00      & 0.00 & 0.00 \\ 
mbert     & 0.30      & 0.36 & 0.37 \\ 
\bottomrule
\end{tabular}
\caption{F1-scores of the two classification tasks: Bangla zero-shot baseline 
and with $k$ retrieval augmented prompts.}
\label{tab:results of classification test}
\end{table}

\begin{table*}[]
\footnotesize
\centering \begin{tabular}{lccccccccc}
\toprule
\multicolumn{1}{l}{}                 
& \multicolumn{3}{c}{zero shot}                        
& \multicolumn{3}{c}{k=1}                          
& \multicolumn{3}{c}{k=3}      
\\ \midrule
\multicolumn{1}{l}{bloomz-3b}                 
& \multicolumn{1}{c}{precision} 
& \multicolumn{1}{c}{recall}
& \multicolumn{1}{c}{f1-score}   
& \multicolumn{1}{c}{precision}
& \multicolumn{1}{c}{recall} 
& \multicolumn{1}{c}{f1-score}
& \multicolumn{1}{c}{precision}
& \multicolumn{1}{c}{recall} 
& \multicolumn{1}{c}{f1-score}    
\\ \midrule
\multicolumn{1}{l}{accuracy}        
& \multicolumn{1}{l}{}         
& \multicolumn{1}{l}{}      
& \multicolumn{1}{c}{{0.33}}
& \multicolumn{1}{l}{}      
& \multicolumn{1}{l}{}       
& \multicolumn{1}{c}{0.35} 
& \multicolumn{1}{l}{}     
& \multicolumn{1}{l}{}      
& \multicolumn{1}{c}{\textbf{0.36}} 
\\ 
\multicolumn{1}{l}{macro avg}      
& \multicolumn{1}{c}{0.15}     
& \multicolumn{1}{c}{0.33} 
& \multicolumn{1}{c}{\textbf{0.20}}    
& \multicolumn{1}{c}{0.18}  
& \multicolumn{1}{c}{0.34} 
& \multicolumn{1}{c}{\textbf{0.20}} 
& \multicolumn{1}{c}{0.26} 
& \multicolumn{1}{c}{0.26}  
& \multicolumn{1}{c}{{0.17}}
\\ 
\multicolumn{1}{l}{weighted avg}    
& \multicolumn{1}{c}{0.14}      
& \multicolumn{1}{c}{0.33}  
& \multicolumn{1}{c}{{0.19}}
& \multicolumn{1}{c}{0.15}    
& \multicolumn{1}{c}{0.35}  
& \multicolumn{1}{c}{{0.20}}   
& \multicolumn{1}{c}{0.42}    
& \multicolumn{1}{c}{0.36}  
& \multicolumn{1}{c}{\textbf{0.24}} 
\\ \toprule
\multicolumn{1}{l}{mbert}                 
& \multicolumn{1}{c}{precision} 
& \multicolumn{1}{c}{recall}
& \multicolumn{1}{c}{f1-score}   
& \multicolumn{1}{c}{precision}
& \multicolumn{1}{c}{recall} 
& \multicolumn{1}{c}{f1-score}
& \multicolumn{1}{c}{precision}
& \multicolumn{1}{c}{recall} 
& \multicolumn{1}{c}{f1-score}    
\\ \midrule                       
\multicolumn{1}{l}{accuracy}     
& \multicolumn{1}{l}{}      
& \multicolumn{1}{l}{}     
& \multicolumn{1}{c}{0.22} 
& \multicolumn{1}{l}{}      
& \multicolumn{1}{l}{}      
& \multicolumn{1}{c}{0.32} 
& \multicolumn{1}{l}{}      
& \multicolumn{1}{l}{}      
& \multicolumn{1}{c}{\textbf{0.33}}
\\ 
\multicolumn{1}{l}{macro avg} 
& \multicolumn{1}{c}{0.31}   
& \multicolumn{1}{c}{0.30}  
& \multicolumn{1}{c}{0.18}   
& \multicolumn{1}{c}{0.52} 
& \multicolumn{1}{c}{0.29} 
& \multicolumn{1}{c}{\textbf{0.21}} 
& \multicolumn{1}{c}{0.18} 
& \multicolumn{1}{c}{0.28}  
& \multicolumn{1}{c}{\textbf{0.21}}
\\ 
\multicolumn{1}{l}{weighted avg}  
& \multicolumn{1}{c}{0.40}    
& \multicolumn{1}{c}{0.22}
& \multicolumn{1}{c}{0.21} 
& \multicolumn{1}{c}{0.62} 
& \multicolumn{1}{c}{0.32} 
& \multicolumn{1}{c}{0.28}
& \multicolumn{1}{c}{0.26} 
& \multicolumn{1}{c}{0.33}  
& \multicolumn{1}{c}{\textbf{0.29}} 
\\ \bottomrule
\end{tabular}
\\
\begin{tabular}{lccccccccc}
\toprule
\multicolumn{1}{l}{}             
& \multicolumn{3}{c}{zero shot} 
& \multicolumn{3}{c}{k=1}
& \multicolumn{3}{c}{k=3}\\
\midrule
\multicolumn{1}{l}{bloomz-3b}             
& \multicolumn{1}{c}{precision} 
& \multicolumn{1}{c}{recall} 
& \multicolumn{1}{c}{f1-score}      
& \multicolumn{1}{c}{precision} 
& \multicolumn{1}{c}{recall} 
& \multicolumn{1}{c}{f1-score} 
& \multicolumn{1}{c}{precision} 
& \multicolumn{1}{c}{recall} 
& \multicolumn{1}{c}{f1-score} 
\\ 
\midrule
\multicolumn{1}{l}{accuracy}     
& \multicolumn{1}{l}{}          
& \multicolumn{1}{l}{}       
& \multicolumn{1}{c}{\textbf{0.61}}
& \multicolumn{1}{l}{}          
& \multicolumn{1}{l}{}       
& \multicolumn{1}{c}{0.60}     
& \multicolumn{1}{l}{}          
& \multicolumn{1}{l}{}       
& \multicolumn{1}{c}{\textbf{0.61}}\\ 
\multicolumn{1}{l}{macro avg}    
& \multicolumn{1}{c}{0.31}      
& \multicolumn{1}{c}{0.37}   
& \multicolumn{1}{c}{0.34}          
& \multicolumn{1}{c}{0.48}      
& \multicolumn{1}{c}{0.48}   
& \multicolumn{1}{c}{\textbf{0.44}}    
& \multicolumn{1}{c}{0.47}      
& \multicolumn{1}{c}{0.48}   
& \multicolumn{1}{c}{\textbf{0.44}} \\
\multicolumn{1}{l}{weighted avg} 
& \multicolumn{1}{c}{0.51}      
& \multicolumn{1}{c}{0.61}   
& \multicolumn{1}{c}{\textbf{0.55}} 
& \multicolumn{1}{c}{0.53}      
& \multicolumn{1}{c}{0.60}   
& \multicolumn{1}{c}{0.54}    
& \multicolumn{1}{c}{0.53}     
& \multicolumn{1}{c}{0.61}  
& \multicolumn{1}{c}{0.54}\\ 
\midrule
\multicolumn{1}{l}{mbert}             
& \multicolumn{1}{c}{precision} 
& \multicolumn{1}{c}{recall} 
& \multicolumn{1}{c}{f1-score}
& \multicolumn{1}{c}{precision} 
& \multicolumn{1}{c}{recall} 
& \multicolumn{1}{c}{f1-score} 
& \multicolumn{1}{c}{precision} 
& \multicolumn{1}{c}{recall} 
& \multicolumn{1}{c}{f1-score} \\ 
\midrule
\multicolumn{1}{l}{accuracy}     
& \multicolumn{1}{l}{}        
& \multicolumn{1}{l}{}     
& \multicolumn{1}{c}{0.35} 
& \multicolumn{1}{l}{}    
& \multicolumn{1}{l}{}     
& \multicolumn{1}{c}{0.37} 
& \multicolumn{1}{l}{}     
& \multicolumn{1}{l}{}     
& \multicolumn{1}{c}{\textbf{0.39}}
\\ 
\multicolumn{1}{l}{macro avg}  
& \multicolumn{1}{c}{0.38}  
& \multicolumn{1}{c}{0.34} 
& \multicolumn{1}{c}{0.30} 
& \multicolumn{1}{c}{0.40} 
& \multicolumn{1}{c}{0.38} 
& \multicolumn{1}{c}{0.36} 
& \multicolumn{1}{c}{0.42} 
& \multicolumn{1}{c}{0.39}  
& \multicolumn{1}{c}{\textbf{0.37}}
\\ 
\multicolumn{1}{l}{weighted avg}
& \multicolumn{1}{c}{0.43}  
& \multicolumn{1}{c}{0.35}  
& \multicolumn{1}{c}{0.34}  
& \multicolumn{1}{c}{0.47}  
& \multicolumn{1}{c}{0.37}
& \multicolumn{1}{c}{0.39}
& \multicolumn{1}{c}{0.48} 
& \multicolumn{1}{c}{0.39}  
& \multicolumn{1}{c}{\textbf{0.41}}
\\ 
\bottomrule
\end{tabular}

\caption{Confusion matrix of main method in Vio-Lens (top) and SentNoB (bottom) test set of BLOOMZ-3b and mBERT.}
\label{tab:confusion matrix of main method in Vio-Lens test of bloomz-3b and mbert}
\end{table*}

Table~\ref{tab:results of classification test} provides an overview of the results of classification. With the instructions of $k=3$ retrieval augmented English prompts, we enhance the F1-scores of Bloomz-3b on the two tasks by 5\% and 10\% respectively.
While Bloom-3b, without instruction tuning compared to Bloomz-3b, cannot generate any meaningful result, suggesting that instruction tuning has a strong impact on retrieval augmented in-context learning.
The traditional masked MLM, mBERT, also gained improvement by 8\% and 7\%.

To facilitate a comprehensive understanding of the performance and discrepancies associated with each task, we present confusion matrices for analysis as follows.
Given the confusion matrix in Table~\ref{tab:confusion matrix of main method in Vio-Lens test of bloomz-3b and mbert} , we find that:

1) With a general assessment across micro, macro, and weighted F1 scores,  Bloomz-3b and mBERT gained improvement from the retrieval prompts.
2) Compare the two models, Bloomz-3b's zero-shot setting tends to misclassify ``non-violence'' and ``Neutral'', and has a reduced macro F1 compared to its weighted F1, while mBERT has a more balanced distribution of confusion between ``non-violence'' (``Neutral'') and the other classes.
This may indicate that for classification tasks, the text generation struggles more with minority classes compared to masked prediction.

\subsection{Results of summarisation task}
The Table~\ref{tab:results of summarisation task} compares several models and methods for summarization task.

\begin{table}[]
\footnotesize
\centering \begin{tabular}{ccccc}
\toprule
\multicolumn{1}{l}{} & R-1 & R-2 & R-L & R-LSum \\ 
\midrule
LEAD-64                & 18.17  & 5.23   & 12.73  & 12.74     \\ 
\midrule
zero shot&&&&\\
mt5-base      & 5.01   & 0.84   & 4.83   & 4.84      \\ 
bloomz-1b1   & 22.08  & 7.11   & 18.43  & 18.44     \\ 
bloomz-3b    & 22.36  & 7.88   & 18.60  & 18.58     \\ 
\midrule
k=1&&&&\\
mt5-base          & 0.97   & 0.13   & 0.91   & 0.92      \\ 
bloomz-1b1        & 10.84  & 2.80   & 9.11   & 9.12      \\ 
bloomz-3b         & 6.61   & 1.52   & 5.56   & 5.55      \\ 
\bottomrule
\end{tabular}
\caption{Rouge scores of Bangla summarization.}
\label{tab:results of summarisation task}
\end{table}

\paragraph{LEAD-64} As an extractive method, it performs well across all metrics. This indicates that in many cases the first few sentences or tokens of an article or document provide a fairly informative summary. As expected, LEAD-64 outperforms the mt5 base model in the zero-shot setting, but is outperformed by the Bloomz models in the same scenario.

\paragraph{Zero-Shot Models}
mt5-base produces the lowest scores across all metrics, suggesting that it struggles to produce satisfactory summaries without domain-specific fine-tuning or data augmentation.
Both bloomz-1b1 and bloomz-3b show significantly better performance, with bloomz-3b having a slight edge over bloomz-1b1, especially in bigram capture (R-2).

\paragraph{Retrieval augmentation with k=1}
Retrieval augmentation seems to drastically affect the performance of mt5-base, reducing its score considerably. This could be due to noise introduced by the retrieved sample or ineffective use of the additional information.
For the Bloomz models, bloomz-1b1 still retains decent performance, although there's a drop when compared to its zero-shot performance. Surprisingly, blommz-3b shows a sharper drop, suggesting that the additional retrieval data may be more of a distraction than an advantage for this model configuration in the summarization task.

\begin{figure*}
    \centering
    \includegraphics[width=0.49\linewidth]{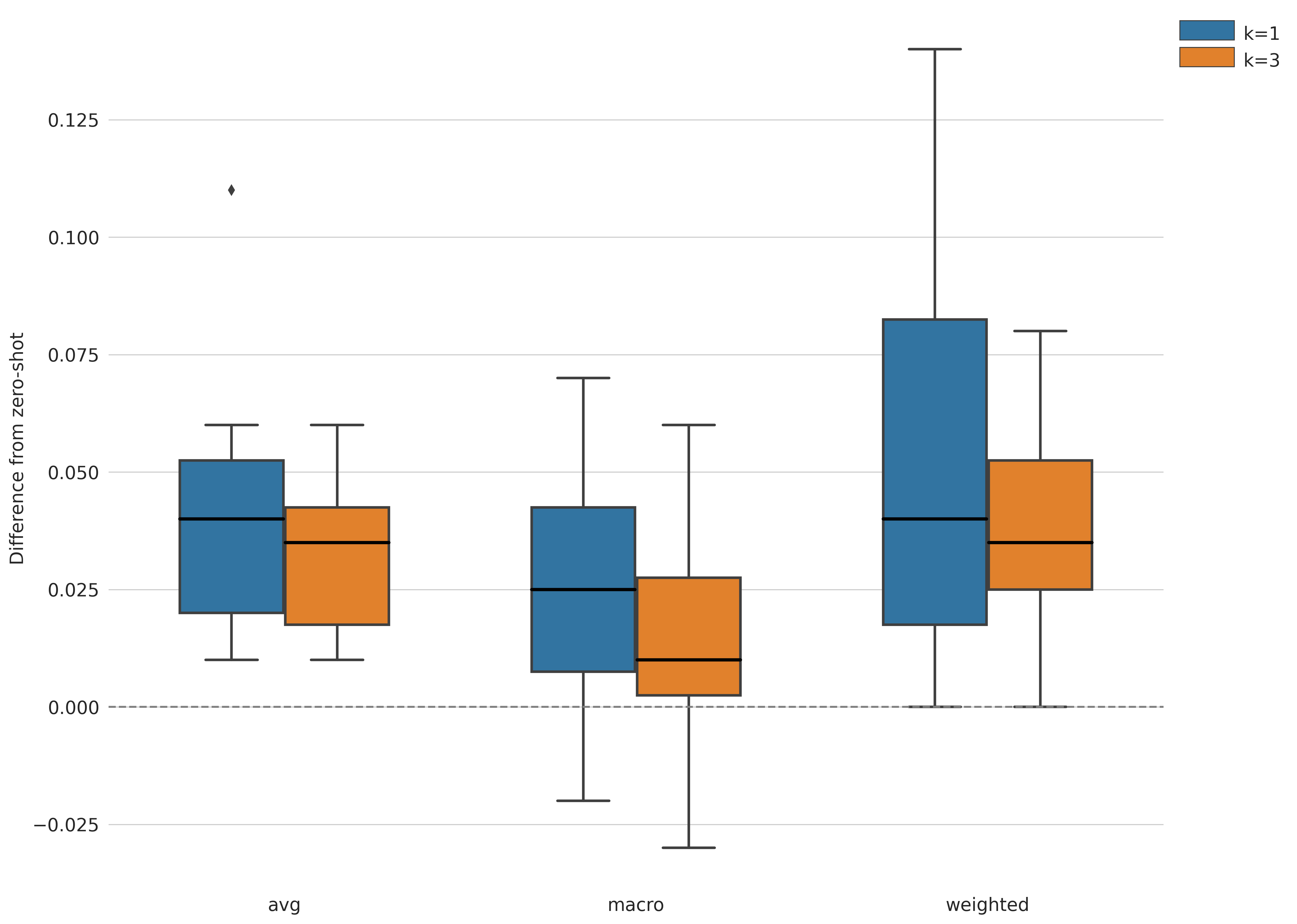}
    \includegraphics[width=0.49\linewidth]{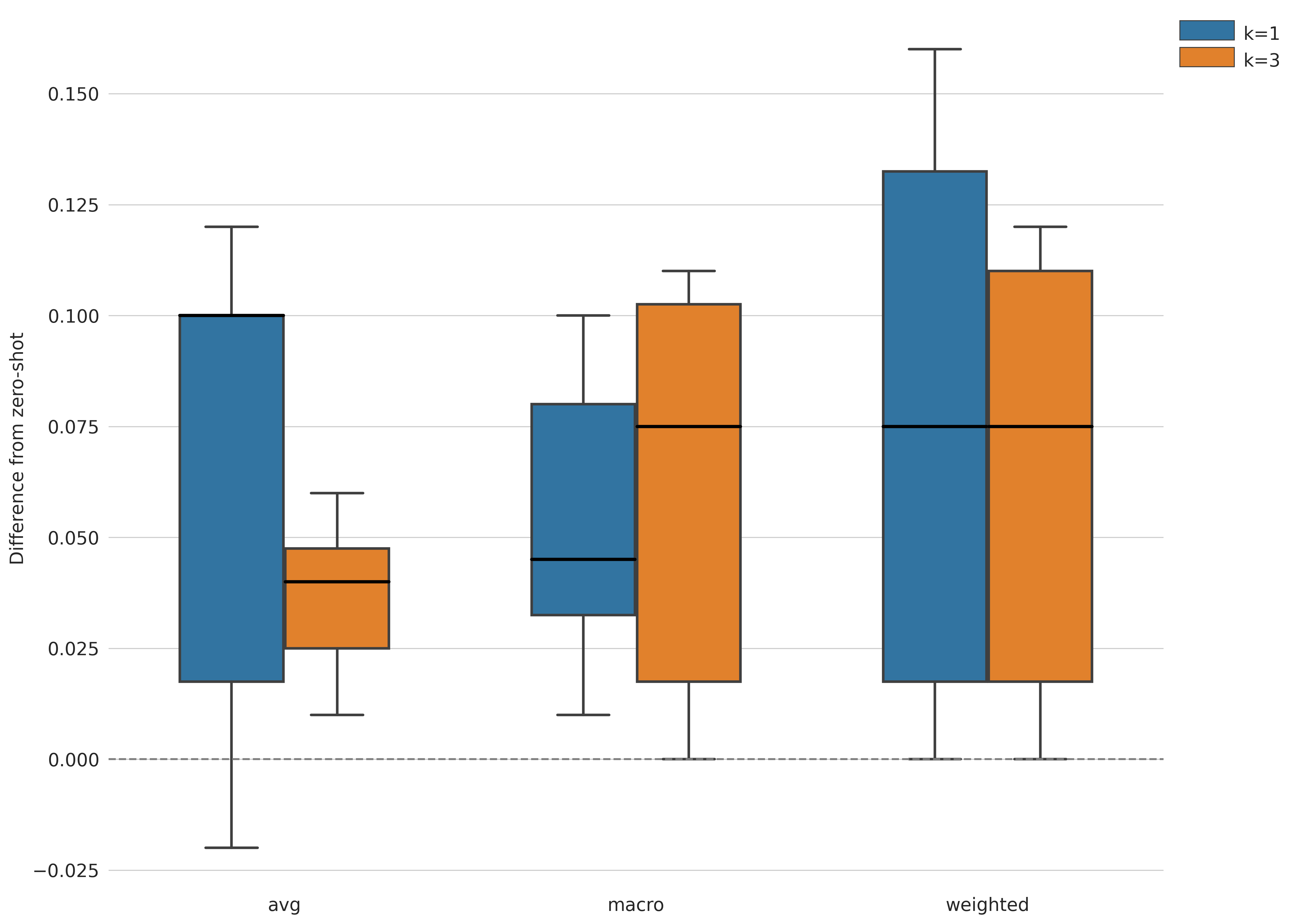}
    \caption{Model performance over differences between zero-shot (represented as `0' on the y-axis) and main method with k=1 and k=3 demonstrations for Vio-Lens test set using bloomz-3b (left) and mbert (right).The y-axis shows the deviations of the main method from the zero-shot values. The statistics are based on 8 and 6 templates, shown in Appendix Table~\ref{tab:8_templates_bloomz3b_task1} and Table~\ref{tab:6_templates_mbert_task1}, respectively.}
    \label{fig:boxplot_task1_test_bloomz3b}
\end{figure*}

\subsection{Analysis and Discussion}

When examining the performance of different models on different tasks, several key observations emerge that are related to linguistic nuances, the underlying language models, and resource allocation.

For classification tasks, it's clear that models with a strong grasp of complex sentence structure and deeper semantics, such as the Bloomz-3b, are more adept at distinguishing nuanced categories like ``passive violence'' or the more ambiguous ``neutral'' sentiment. This aptitude likely stems from their ability to understand context better than their simpler counterparts. In parallel, the critical role of zero-shot learning becomes apparent. The ability of a model to generalize a task without specific fine-tuning speaks volumes about its robustness. For example, in our studies, models such as the Bloomz-3b showed commendable performance in a zero-shot setting. Furthermore, as we played around with the variable k (representing the number of samples retrieved), it was instructive to see that a larger value didn't always translate into better performance. This underscores the nuanced ability of a model to sift through information and potentially eliminate noise.

Turning to the summarization task, coherence and relevance seem to be the pillars of excellence. Advanced models are more adept at weaving sentences that are not only structurally coherent, but also rich in information. This finesse is evident in the superior Rouge scores of the models. The dichotomy between generative and extractive approaches is also evident. While generative models, including mt5-base and Bloomz-1b1, outperformed the extractive model (LEAD-64) in a zero-shot framework, they seemed a bit sensitive when retrieval augmentation came into play.

Finally, when it comes to resource distribution, there's an undeniable correlation between performance and computational resources. The stellar performance of models like Bloomz-3b likely comes at the cost of intense computational demands. However, one must consider the cost-benefit ratio. In addition, the drop in performance of these models with retrieval augmentation at k=1 suggests a potential sensitivity to the balance or diversity of the dataset.

For the summarization task, an interesting observation is that more extensive models don't always outperform on all metrics, suggesting that we need to be more discriminating in our resource allocation. The significant performance drop with retrieval augmentation further supports this argument.

To conclude this analysis, while modern language models are capable of handling complex tasks, they require careful configuration and thoughtful resource distribution. Unraveling the complexity of these models can pave the way for optimized solutions in both classification and summarization.
\section{Ablation Study}

\subsection{The Stability across Templates}

In our experiment for Vio-Lens, we compared the performance of Bloomz-3b and mbert, in terms of their ability to classify text samples into categories. In order to assess the effectiveness of the retrieval augmented prompting method compared to the zero-shot baseline, we conduct a statistic across different templates.

For Bloomz-3b and mBERT, we test different prompt templates 
, and created a boxplot (Figure~\ref{fig:boxplot_task1_test_bloomz3b}) to visualize the difference of F1 scores from our main method to the zero-shot baseline across templates.
It's shown that with the retrieval augmented English prompts under different templates, both models achieved a stable improvement compared to the Bangla zeroshot baseline. Aso it's clear that mBERT, on average, shows greater improvements in F1 scores when transitioning from the zero-shot baseline to retrieval augmented prompting, compared to Bloomz-3b. 

\subsection{Impact of Bangla and Hindi Prompt Template}
\begin{table*}[]
\footnotesize
\centering \begin{tabular}{lccccccc}
\toprule
& \multicolumn{3}{c}{k=1}        
& \multicolumn{3}{c}{k=3}  
\\ 
& \multicolumn{1}{c}{precision} 
& \multicolumn{1}{c}{recall} 
& \multicolumn{1}{c}{f1-score} 
& \multicolumn{1}{c}{precision} 
& \multicolumn{1}{c}{recall} 
& \multicolumn{1}{c}{f1-score} 
\\ \midrule
bangla prompt & \multicolumn{7}{l}{
{\bng paThY}: \{text\} {\bng inm/nilikht ibkl/pguil ed{O}Ja paeThYr jnY sm/bhabY Anubhuuit kii?}
}\\ 

accuracy    
& \multicolumn{1}{c}{} 
& \multicolumn{1}{c}{}  
& \multicolumn{1}{c}{0.14}  
& \multicolumn{1}{c}{}    
& \multicolumn{1}{c}{}      
& \multicolumn{1}{c}{0.45} 
\\
macro avg  
& \multicolumn{1}{c}{0.34}   
& \multicolumn{1}{c}{0.09}   
& \multicolumn{1}{c}{0.13}  
& \multicolumn{1}{c}{0.32} 
& \multicolumn{1}{c}{0.28} 
& \multicolumn{1}{c}{0.29}
\\
weighted avg  
& \multicolumn{1}{c}{0.51}  
& \multicolumn{1}{c}{0.14}
& \multicolumn{1}{c}{0.21}  
& \multicolumn{1}{c}{0.49} 
& \multicolumn{1}{c}{0.45} 
& \multicolumn{1}{c}{0.46} 
\\ \midrule
hindi prompt  & \multicolumn{7}{c}{
{\dn pAW}: \{text\} {\dn En\3DFwElEKt EvkSpo{\qva} ko d\?Kt\? \7{h}e pAW k\? Ele s\2BAEvt BAvnA \3C8wA h\4{\rs ?\re}}
}\\ 
accuracy     
& \multicolumn{1}{c}{}       
& \multicolumn{1}{c}{}       
& \multicolumn{1}{c}{0.39}   
& \multicolumn{1}{c}{}      
& \multicolumn{1}{c}{}      
& \multicolumn{1}{c}{0.54}   
\\
macro avg   
& \multicolumn{1}{c}{0.34}    
& \multicolumn{1}{c}{0.28}  
& \multicolumn{1}{c}{0.29}  
& \multicolumn{1}{c}{0.34}  
& \multicolumn{1}{c}{0.34}  
& \multicolumn{1}{c}{0.34}  
\\
weighted avg
& \multicolumn{1}{c}{0.51}  
& \multicolumn{1}{c}{0.39}  
& \multicolumn{1}{c}{0.43}  
& \multicolumn{1}{c}{0.52}  
& \multicolumn{1}{c}{0.54}  
& \multicolumn{1}{c}{0.53} 
\\
\bottomrule
\end{tabular}
\caption{Results of prompt template in bangla and hindi of main method in SentNoB test of bloomz-3b.}
\label{tab:results of SentNoB test bloomz-3b}
\end{table*}

Instead of English, we further explore applying Bangla itself and its linguistically similar high-resource language Hindi as the language of the prompt template 
, as shown in Table \ref{tab:results of SentNoB test bloomz-3b}.

Main method with English prompt: This configuration yields the highest macro average F1 score of all three prompt templates. 

Hindi Prompt Template: While the Hindi prompt template leads to significant improvements in precision and recall for individual categories such as ``Neutral'', the macro average F1 score is still lower than that of the main method with the English prompt.

Bangla prompt template: The Bangla prompt template, while showing some improvements in precision for specific categories such as ``positive'', experiences a decrease in recall and overall accuracy. As a result, the macro average F1 score is the lowest of the three templates.

This means that while the Bangla prompt template may improve performance for specific categories, it has an overall negative impact on the model's ability to generalize across all categories in the SentNoB test. Conversely, the Hindi prompt template's improvements in precision and recall for individual categories don't translate into a higher macro average F1 score compared to the main method with the English prompt.

In summary, the macro average F1-score results show that the main method with the English prompt template remains the most effective overall. However, the choice of prompt template can significantly affect performance for specific categories, as demonstrated by the Hindi and Bangla templates. This nuanced understanding underscores the need to balance category-specific and overall performance when selecting prompt templates in cross-lingual retrieval augmentation.

\subsection{Impact of Hindi sentence pool}

Comparing the results in Table \ref{tab:task2 test hindi dataset} with the previous experiments, we observe that the Hindi retrieval dataset generally improves the model's ability to retrieve ``Neutral'' content in the mBERT model. However, the model continues to struggle with the ``Neutral'' category, with low recall and F1 scores, regardless of the sentence pool used. This suggests that further refinements may be needed to improve retrieval accuracy for neutral sentiment sentences.
The studies with Hindi retrieval data show that both bloomz-3b and mbert don't show any improvements compared to the main method with the English prompt template. This suggests that while using alternative retrieval datasets can improve performance for specific sentiment categories, the choice of retrieval data may need to be carefully considered to maximize overall performance across categories in cross-lingual sentiment analysis tasks.

\section{Conclusion}
In this paper, we have introduced a novel approach to address the challenges of applying Large Language Models to low-resource languages, with a focus on Bangla. Our methodology employs cross-lingual retrieval-augmented in-context learning, thereby enriching the capabilities of MPLMs, specifically BLOOM and BLOOMZ. We have extensively tested our approach on two classification tasks and one summarization task.

Our experimental results demonstrate the effectiveness of our approach in achieve superior F1 scores for classification tasks. 

Upon further analysis, the cross-lingual retrieval mechanism contributes significantly to the model's performance.

This work lays the foundation for further studies on the application of cross-lingual retrieval and in-context learning methods in low-resource languages. Future work could extend this approach to even more underrepresented languages and potentially adapt it to more complex NLP tasks such as question answering or machine translation.

\section*{Limitations}
While our study has yielded promising results, it is not without limitations. The effectiveness of retrieval augmentation is also tied to the model architecture, and its impact on different models remains largely unexplored. In addition, the availability of specific language datasets for sentence retrieval and resource constraints remain practical challenges. Further exploration of prompt design and consideration of external factors could improve our methodology. Acknowledging these limitations is essential for a full interpretation of our results and the direction of future research.

\section*{Acknowledgements}
This work was supported by Leibniz Supercomputing Centre (LRZ), Munich Center for Machine
Learning (MCML) and China Scholarship Council (CSC).

\bibliography{anthology,custom}
\bibliographystyle{acl_natbib}

\appendix

\section{Appendix}
\label{sec:appendix}

\begin{table*}[]
\footnotesize
\centering
\begin{tabular}{lccc}
\toprule
             & zero-shot                                                               & k=1                                                                & k=3                                                             \\
\midrule
prompt       & \multicolumn{3}{l}{\{text\} Direct Aggression, Indirect Aggression, or No Aggression?}                                                                                                                         \\
accuracy     & 0.53                                                                    & 0.54                                                               & 0.54                                                            \\
macro avg    & 0.17                                                                    & 0.18                                                               & 0.18                                                            \\
weighted avg & 0.38                                                                    & 0.38                                                               & 0.38                                                            \\
\midrule
prompt       & \multicolumn{3}{l}{\begin{tabular}[c]{@{}l@{}}Evaluate the text: ’\{text\}’. Would you categorize it as absence of aggression, mild aggression,\\ or strong aggression?\end{tabular}}                          \\
accuracy     & 0.18                                                                    & 0.23                                                               & 0.23                                                            \\
macro avg    & 0.17                                                                    & 0.15                                                               & 0.15                                                            \\
weighted avg & 0.13                                                                    & 0.16                                                               & 0.16                                                            \\
\midrule
prompt       & \multicolumn{3}{l}{\begin{tabular}[c]{@{}l@{}}In the context of ’\{text\}’, which category best captures its aggression level: absence of aggression,\\ mild aggression, or strong aggression?\end{tabular}}   \\
accuracy     & 0.12                                                                    & 0.15                                                               & 0.16                                                            \\
macro avg    & 0.1                                                                     & 0.15                                                               & 0.16                                                            \\
weighted avg & 0.06                                                                    & 0.11                                                               & 0.12                                                            \\
\midrule
prompt       & \multicolumn{3}{l}{\begin{tabular}[c]{@{}l@{}}For the text: ’\{text\}’, ascertain its aggression scale: absence of aggression, mild aggression, or\\ strong aggression?\end{tabular}}                          \\
accuracy     & 0.19                                                                    & 0.21                                                               & 0.21                                                            \\
macro avg    & 0.13                                                                    & 0.14                                                               & 0.14                                                            \\
weighted avg & 0.14                                                                    & 0.16                                                               & 0.15                                                            \\
\midrule
prompt       & \multicolumn{3}{l}{\begin{tabular}[c]{@{}l@{}}From the following choices, which resonates with the theme of ’\{text\}’? Options: No Intensity,\\ Low Intensity, High Intensity\end{tabular}}                   \\
accuracy     & 0.13                                                                    & 0.24                                                               & 0.19                                                            \\
macro avg    & 0.1                                                                     & 0.17                                                               & 0.15                                                            \\
weighted avg & 0.12                                                                    & 0.26                                                               & 0.2                                                             \\
\midrule
prompt       & \multicolumn{3}{l}{\begin{tabular}[c]{@{}l@{}}From the following choices, which resonates with the theme of ’\{text\}’? Options: no intensity,\\ low intensity, high intensity\end{tabular}}                   \\
accuracy     & 0.23                                                                    & 0.28                                                               & 0.27                                                            \\
macro avg    & 0.18                                                                    & 0.22                                                               & 0.2                                                             \\
weighted avg & 0.22                                                                    & 0.31                                                               & 0.26                                                            \\
\midrule
prompt       & \multicolumn{3}{l}{\begin{tabular}[c]{@{}l@{}}In the context of the text ’\{text\}’, which of the following best describes its tone? Options: No\\ Intensity, Low Intensity, High Intensity\end{tabular}}      \\
accuracy     & 0.14                                                                    & 0.2                                                                & 0.15                                                            \\
macro avg    & 0.11                                                                    & 0.15                                                               & 0.12                                                            \\
weighted avg & 0.1                                                                     & 0.18                                                               & 0.13                                                            \\
\midrule
prompt       & \multicolumn{3}{l}{\begin{tabular}[c]{@{}l@{}}Reflecting on the statement ’\{text\}’, which aggressive level does it resonate with: non-aggressive,\\ slightly aggressive, or highly aggressive?\end{tabular}} \\
accuracy     & 0.33                                                                    & 0.35                                                               & 0.36                                                            \\
macro avg    & 0.2                                                                     & 0.2                                                                & 0.17                                                            \\
weighted avg & 0.19                                                                    & 0.2                                                                & 0.24     \\                   \bottomrule                                  
\end{tabular}
\caption{F1-score results with 8 prompt templates of Vio-Lens test using bloomz-3b model}
\label{tab:8_templates_bloomz3b_task1}
\end{table*}

\begin{table*}[]
\centering 
\begin{tabular}{lccc}
\toprule
             & zero-shot                       & k=1                      & k=3                     \\
\midrule
prompt       & \multicolumn{3}{l}{The text displays {[}MASK{]} aggression: \{text\}}                \\
verbalizer   & \multicolumn{3}{l}{direct, indirect, none}                                           \\
accuracy     & 0.36                            & 0.35                     & 0.36                    \\
macro avg    & 0.22                            & 0.23                     & 0.23                    \\
weighted avg & 0.31                            & 0.31                     & 0.31                    \\
\midrule
prompt       & \multicolumn{3}{l}{Considering aggressive tendencies, this is {[}MASK{]}: \{text\}}  \\
verbalizer   & \multicolumn{3}{l}{overt, covert, absent}                                            \\
accuracy     & 0.1                             & 0.2                      & 0.17                    \\
macro avg    & 0.07                            & 0.17                     & 0.14                    \\
weighted avg & 0.03                            & 0.19                     & 0.15                    \\
\midrule
prompt       & \multicolumn{3}{l}{From an aggression perspective, the text is {[}MASK{]}: \{text\}} \\
verbalizer   & \multicolumn{3}{l}{overt, covert, absent}                                            \\
accuracy     & 0.12                            & 0.22                     & 0.2                     \\
macro avg    & 0.09                            & 0.18                     & 0.16                    \\
weighted avg & 0.06                            & 0.21                     & 0.18                    \\
\midrule
prompt       & \multicolumn{3}{l}{The described behavior in \{text\} is {[}MASK{]} aggression.}     \\
verbalizer   & \multicolumn{3}{l}{explicit, implicit, neutral}                                      \\
accuracy     & 0.24                            & 0.36                     & 0.35                    \\
macro avg    & 0.19                            & 0.24                     & 0.23                    \\
weighted avg & 0.23                            & 0.31                     & 0.3                     \\
\midrule
prompt       & \multicolumn{3}{l}{The underlying theme in \{text\} is {[}MASK{]} aggression.}       \\
verbalizer   & \multicolumn{3}{l}{assaultive, indirect, peaceful}                                   \\
accuracy     & 0.22                            & 0.32                     & 0.33                    \\
macro avg    & 0.18                            & 0.21                     & 0.21                    \\
weighted avg & 0.21                            & 0.28                     & 0.29                    \\
\midrule
prompt       & \multicolumn{3}{l}{\{text\} is interpreted as {[}MASK{]} aggression.}                \\
verbalizer   & \multicolumn{3}{l}{assaultive, indirect, peaceful}                                   \\
accuracy     & 0.51                            & 0.49                     & 0.51                    \\
macro avg    & 0.23                            & 0.27                     & 0.25                    \\
weighted avg & 0.37                            & 0.37                     & 0.37                   \\
\bottomrule
\end{tabular}
\caption{F1-score results with 6 prompt templates of Vio-Lens test using mBert model}
\label{tab:6_templates_mbert_task1}
\end{table*}

\begin{table*}[]
\centering
\begin{tabular}{ccccccc}
\toprule
             & \multicolumn{3}{c}{k=1}       & \multicolumn{3}{c}{k=3}       \\ \midrule
bloomz-3b    & precision & recall & f1-score & precision & recall & f1-score \\ \midrule
Negative     & 0.58      & 0.84   & 0.69     & 0.59      & 0.88   & 0.70     \\
Neutral      & 0.09      & 0.00   & 0.00     & 0.08      & 0.00   & 0.00     \\
Positive     & 0.55      & 0.49   & 0.52     & 0.58      & 0.47   & 0.52     \\
accuracy     &           &        & 0.57     &           &        & 0.58     \\
macro avg    & 0.41      & 0.44   & 0.40     & 0.42      & 0.45   & 0.41     \\
weighted avg & 0.48      & 0.57   & 0.51     & 0.49      & 0.58   & 0.51     \\ \toprule
mbert        & precision & recall & f1-score & precision & recall & f1-score \\ \midrule
Negative     & 0.48      & 0.24   & 0.32     & 0.48      & 0.33   & 0.39     \\
Neutral      & 0.21      & 0.34   & 0.26     & 0.21      & 0.28   & 0.24     \\
Positive     & 0.27      & 0.37   & 0.31     & 0.25      & 0.33   & 0.28     \\
accuracy     &           &        & 0.30     &           &        & 0.32     \\
macro avg    & 0.32      & 0.32   & 0.30     & 0.31      & 0.31   & 0.31     \\
weighted avg & 0.36      & 0.30   & 0.30     & 0.36      & 0.32   & 0.33     \\ \bottomrule
\end{tabular}

\caption{Results in SentNoB test of BLOOMZ-3b and mBERT with hindi retrieval corpus.}
\label{tab:task2 test hindi dataset}
\end{table*}


\end{document}